\documentclass[12pt]{article}

\usepackage{graphicx}
\usepackage{amsmath,amssymb,amsfonts}
\usepackage{algpseudocode,algorithm}
\usepackage{subcaption}
\usepackage{makecell}

\usepackage{times}
\usepackage{color}
\usepackage{multirow}
\usepackage[authoryear]{natbib}
\usepackage{rotating}
\usepackage{bbm}
\usepackage{latexsym}

\newcommand{\captionfonts}{\normalsize}

\makeatletter
\long\def\@makecaption#1#2{%
  \vskip\abovecaptionskip
  \sbox\@tempboxa{{\captionfonts #1: #2}}%
  \ifdim \wd\@tempboxa >\hsize
    {\captionfonts #1: #2\par}
  \else
    \hbox to\hsize{\hfil\box\@tempboxa\hfil}%
  \fi
  \vskip\belowcaptionskip}
\makeatother

\begin{document}
\hspace{13.9cm}1

\ \vspace{20mm}\\
{\LARGE Weber-Fechner Law in Temporal Difference learning derived from Control as Inference}

\ \\
{\bf \large Keiichiro Takahashi$^{\displaystyle 1}$, Taisuke Kobayashi$^{\displaystyle 2}$, Tomoya Yamanokuchi$^{\displaystyle 1}$, and Takamitsu Matsubara$^{\displaystyle 1}$}\\
{$^{\displaystyle 1}$Division of Information Science, Nara Institute of Science and Technology, Ikoma, Nara, 630-0192, Japan.}\\
{$^{\displaystyle 2}$National Institute of Informatics (NII) and The Graduate University for Advanced Studies (SOKENDAI),
        2-1-2 Hitotsubashi, Chiyoda-ku, Tokyo, 101-8430, Japan.}\\
%

{\bf Keywords:} Temporal difference learning, Control as inference, Weber-Fechner law

\thispagestyle{empty}
\markboth{}{NC instructions}
\ \vspace{-0mm}\\
\begin{center} {\bf Abstract} \end{center}
This paper investigates a novel nonlinear update rule for value and policy functions based on temporal difference (TD) errors in reinforcement learning (RL).
The update rule in the standard RL states that the TD error is linearly proportional to the degree of updates, treating all rewards equally without no bias.
On the other hand, the recent biological studies revealed that there are nonlinearities in the TD error and the degree of updates, biasing policies optimistic or pessimistic.
Such biases in learning due to nonlinearities are expected to be useful and intentionally leftover features in biological learning.
Therefore, this research explores a theoretical framework that can leverage the nonlinearity between the degree of the update and TD errors.
To this end, we focus on a \textit{control as inference} framework, since it is known as a generalized formulation encompassing various RL and optimal control methods.
In particular, we investigate the uncomputable nonlinear term needed to be approximately excluded in the derivation of the standard RL from control as inference.
By analyzing it, Weber-Fechner law (WFL) is found, namely, perception (a.k.a. the degree of updates) in response to stimulus change (a.k.a. TD error) is attenuated by increase in the stimulus intensity (a.k.a. the value function).
To numerically reveal the utilities of WFL on RL, we then propose a practical implementation using a reward-punishment framework and modifying the definition of optimality.
Analysis of this implementation reveals that two utilities can be expected i) to increase rewards to a certain level early, and ii) to sufficiently suppress punishment.
We finally investigate and discuss the expected utilities through simulations and robot experiments.
As a result, the proposed RL algorithm with WFL shows the expected utilities that accelerate the reward-maximizing startup and continue to suppress punishments during learning.

\section{Introduction}

Reinforcement learning (RL) \citep{sutton2018reinforcement} provides robots with policies that allow them to interact in unknown and complex environments, replacing conventional model-based control with it.
Temporal difference (TD) learning \citep{sutton1988learning} is a fundamental methodology in RL.
For example, it has been introduced as the basis for proximal poicy optimization (PPO) \citep{schulman2017proximal} and soft actor-critic (SAC) \citep{haarnoja2018soft}, the most famous algorithms in the the recent years, both of which are implemented on popular RL libraries \citep{raffin2021stable,huang2022cleanrl},
and applied to many real robots \citep{andrychowicz2020learning,wahid2021learning,nematollahi2022robot,kaufmann2023champion,radosavovic2024real}.
In TD learning, the future value predicted from the current state is compared to the one from the state after transition, which is so-called TD error.
The value function for that prediction can be learned by making this TD error zero, and its learning convergence is theoretically supported by Bellman equation (although some residuals tend to remain in practice).
In addition, actor-critic methods often utilize the TD error as the weight of policy gradient, since it indicates the direction of maximizing the future value.

Although TD learning plays an important role in RL theories and algorithms as above, TD learning can explain many biological behaviors.
That is, a strong correlation between TD errors and the amount of dopamine or the firing rate of dopamine neurons, which affect memory and learning in organisms, has been reported \citep{schultz1993responses,o2003temporal,starkweather2021dopamine}, and behavioral learning in organisms is also hypothesized to be based on RL \citep{dayan2002reward,doya2021canonical}.
Recently, more detailed investigation of the relationship between TD errors and dopamines has revealed that it is not a simple linear relationship as suggested by the standard TD learning, but is biased and nonlinear \citep{dabney2020distributional,muller2024distributional}.
It has also been reported that some of nonlinearities may stabilize learning performance \citep{hoxha2024evolving}.
Indeed, in the context of RL theory, nonlinear transformed TD learning has been proposed to obtain risk-sensitive behavior \citep{shen2014risk,noorani2023exponential} and robustness to outliers \citep{sugiyama2009least,cayci2024provably}.
The above studies suggest that the implicit biases introduced by nonlinearities would be effective both theoretically and biologically.
In other words, discovering new nonlinearities theoretically or experimentally and understanding their utilities has both an engineering value, such as robot control, and a biological value, such as modeling the principles of behavioral learning in organisms.

On top of that, \textit{Control as inference} \citep{levine2018reinforcement} is expected to find various nonlinearities because its generalized framework encompasses various RL and optimal control methods.
Indeed, our previous study has found that the conventional TD learning can be approximately derived by giving appropriate definitions of optimality and divergence \citep{kobayashi2022optimistic}.
At the same time, it revealed that updating the value and policy functions according to TD errors becomes optimistic by modifying the definition of the divergence.
In a subsequent study, it was also found that modifying the definition of optimality leads to pessimistic updates \citep{kobayashi2024drop}.
This paper also follows the new derivation of TD learning in these previous studies to find/investigate the novel nonlinearity undiscovered so far.

Specifically, we focus on the fact that an approximation was necessary to derive the conventional TD learning from control as inference with linearity between the TD errors and the degrees of updating.
This approximation was generally unavoidable in order to eliminate an unknown variable and allow numerical computation.
However, as the term excluded by the approximation is nonlinear, it should be worth analyzing its role as the first contribution of this paper.
In addition, given that organisms tend to leave useful utilities in the process of evolution, it is important to consider whether the utilities found analytically are useful in RL in the hope that they are also hidden in organisms.
To numerically evaluate that, we then propose a novel biologically-plausible algorithm that combines a reward-punishment framework \citep{kobayashi2019reward,wang2021modular} with a modified definition of optimality \citep{kobayashi2024drop}, making the nonlinear term computable in any task covered by RL.

As a result, we show analytically that the nonlinear term excluded so far leads to \textit{Weber-Fechner law} (WFL), which is well-known to be biologically-plausible characteristic \citep{scheler2017logarithmic,portugal2011weber,nutter2006role,binhi2023magnetic}.
That is, the degree of update of the value and policy functions corresponding to the intensity of perception is logarithmically affected by the scale of the value function, which is the base stimulus: i.e. with the small scale, the update is sensitive to even a small TD error; and with the large scale, only a large TD error allows the update enough.
This WFL is dominant when the optimality is highly uncertain, while the conventional linear behavior is found when the optimality becomes deterministic.
Although organisms have been reported to behave in a way that reduces the uncertainty of predictions \citep{parr2022active}, paradoxically, they are forced to make decisions under uncertainty \citep{tversky1974judgment}.
Hence, we can anticipate that WFL under the uncertain optimality may be found in the biological relationship between TD errors and dopamines in organisms as well.

Through simulations and real-robot experiments, we also confirm that the RL algorithm with the found WFL can learn the optimal policies properly and exert special effects on learning processes and outcomes.
Specifically, the proposed RL algorithm acquires tasks, and the WFL added in the right balance maximizes rewards eventually while suppressing punishments during learning.
In addition, the capability to accelerate learning from a small reward phase allows the robot to efficiently learn a valve-turning task \citep{ahn2020robel} on a real robot, resulting in a higher task success rate.
Thus, WFL is useful in RL, raising expectations that organisms have same (or similar) utilities.

\section{Preliminaries}

\subsection{Reinforcement learning}

In RL, an aget aims to optimize a learnable policy so that the accumulation of future rewards from an unknown environment (so-called return) is maximized \citep{sutton2018reinforcement} under Markov decision process (MDP).
That is, an environment with a task to be solved is (implicitly) defined as the tuple $(\mathcal{S}, \mathcal{A}, \mathcal{R}, p_0, p_e)$.
Here, $\mathcal{S} \subset \mathbb{R}^{|\mathcal{S}|}$ and $\mathcal{A} \subset \mathbb{R}^{|\mathcal{A}|}$ denote the state and action spaces with the $|\mathcal{S}|$-dimensional state $s$ and the $|\mathcal{A}|$-dimensional action $a$.
$\mathcal{R} \subseteq \mathbb{R}$ is the subset on which rewards exist, and the specific values (and even existences) of its upper and lower boundaries $\mathcal{R} \subseteq (\underline{R}, \overline{R})$ are usually unknown.
$p_0 : \mathcal{S} \mapsto \mathbb{R}_+$ denotes the probability for sampling the initial state of each trajectory, and $p_e : \mathcal{S} \times \mathcal{A} \times \mathcal{S} \mapsto \mathbb{R}_+$ is known as the state transition probability (or dynamics).

With such a definition, the agent repeatedly interacts with the environment at the current state $s$ according to the action $a$ determined by its policy $\pi : \mathcal{S} \times \mathcal{A} \mapsto \mathbb{R}_+$ with its learnable parameters $\phi$, resulting in the next state $s^\prime$ and the corresponding reward $r$, which is computed by the reward function $r : \mathcal{S} \times \mathcal{A} \mapsto \mathcal{R}$.
As a result, the agent obtains the return $R_t$ from the time step $t$ as follows:
\begin{align}
    R_t = (1 - \gamma) \sum_{k=0}^\infty \gamma^k r_{t+k}
    \label{eq:return}
\end{align}
where $\gamma \in [0, 1)$ denotes the discount factor.
Note that $1-\gamma$ is multiplied for normalization to match the implementation used in this paper, although the definition without it is common.

The optimal policy $\pi^\ast$ is defined for this.
\begin{align}
    \pi^\ast(\cdot \mid s) = \arg \max_{\pi} \mathbb{E}_{p_\tau}[R_t \mid s_t = s]
    \label{eq:prob_rl}
\end{align}
where $p_\tau$ denotes the probablity for the trajectory, defined as the joint probability of $p_e$ and $\pi$ from $t$ to $\infty$.
$\phi$ is optimized to represent $\pi^\ast$ for any state.

As a remark, the maximization target is modeled as the (state) value function $V : \mathcal{S} \mapsto \mathcal{R}$ with its learnable parameters $\theta$.
When $a_t = a$ is also given as the additional condition for computing the above expectation as $\mathbb{E}_{p_\tau}[R_t \mid s_t = s, a_t = a]$, the action value function $Q : \mathcal{S} \times \mathcal{A} \mapsto \mathcal{R}$ is defined for its model.
Here, $Q(s)$ can be approximated by $r + \gamma V(s^\prime)$ by following the recursive definition of return, and the difference between it and $V(s)$ is defined as the TD error, $\delta := r + \gamma V(s^\prime) - V(s)$, which should be zero by optimizing $\theta$ for any state.
In addition, $\delta$ can be utilized for updating $\phi$ so that $\pi$ is more likely to generate actions that make $\delta$ more positive (i.e. larger return than expected).

\subsection{Update rule derived from control as inference}

To interpret the above optimal control problem as a kind of inference problem, \textit{control as inference} introduces the stochastic variable for the trajectory's optimality $O = \{0, 1\}$ \citep{levine2018reinforcement}.
As it is relevant to the return, its conditional probability is defined as follows:
\begin{align}
    \begin{split}
        p(O=1 \mid s) &= e^{\beta (V(s) - \overline{R})} =: p_V
        \\
        p(O=1 \mid s, a) &= e^{\beta (Q(s,a) - \overline{R})} =: p_Q
    \end{split}
    \label{eq:prob_optim}
\end{align}
where $\beta \in \mathbb{R}_+$ denotes the inverse temperature parameter
Note that $p(O=0)$ can also be given as $1 - p(O=1)$ since $O$ is binary.
When $\beta$ is small, the optimality is ambiguous, and as getting $\beta$ larger, the optimality becomes deterministic.

With the probability of optimality, the optimal and non-optimal policies are inferred according to Bayes theorem.
Specifically, with the baseline policy $b(a \mid s)$ for sampling actions, $\pi(a \mid s, O)$ is obtained.
\begin{align}
    &\pi(a \mid s, O)
    = \cfrac{p(O \mid s, a) b(a \mid s)}{p(O \mid s)}
    \nonumber \\
    &= \begin{cases}
        \cfrac{e^{\beta (Q(s,a) - \overline{R})}}{e^{\beta (V(s) - \overline{R})}} b(a \mid s) & O = 1
        \\
        \cfrac{1 - e^{\beta (Q(s,a) - \overline{R})}}{1 - e^{\beta (V(s) - \overline{R})}} b(a \mid s) & O = 0
    \end{cases}
    \label{eq:policy_bayes_optim}
\end{align}

Based on this definition, the previous study \citep{kobayashi2022optimistic} has consider the following minimization problem for optimizing $\theta$, the parameters for the value function $V$.
\begin{align}
    \min_\theta \mathbb{E}_{p_e, b}[\mathrm{KL}(p(O \mid s) \mid p(O \mid s, a))]
    \label{eq:prob_val}
\end{align}
where $\mathrm{KL}(p_1 \mid p_2) = \mathbb{E}_{x \sim p_1}[\ln p_1(x) - \ln p_2(x)]$ is Kullback-Leibler divergence.
To solve this problem, its gradient w.r.t. $\theta$, $g_\theta$, is derived as follows:
\begin{align}
    g_\theta &=
    \mathbb{E}_{p_e, b}\Biggl [
    \nabla_\theta p_V \ln \cfrac{p_V}{p_Q}
    - \nabla_\theta p_V \ln \cfrac{1 - p_V}{1 - p_Q}
    + p_V \nabla_\theta \ln p_V
    + (1 - p_V) \nabla_\theta \ln (1 - p_V)
    \Biggr ]
    \nonumber \\
    &= \mathbb{E}_{p_e, b}\Biggl [
    - \nabla_\theta V(s) \beta p_V
    \Biggl \{
    \beta (Q(s,a) - V(s))
    + \ln \cfrac{1 - p_V}{1 - p_Q}
    \Biggr \}
    \Biggr ]
    \nonumber \\
    &\propto \mathbb{E}_{p_e, b}\Biggl [
    - \nabla_\theta V(s)
    \Biggl \{
    (1 - \lambda_\beta) (Q(s,a) - V(s))
    + \lambda_\beta \ln \cfrac{1 - p_V}{1 - p_Q}
    \Biggr \}
    \Biggr ]
    \label{eq:grad_val}
\end{align}
where $\lambda_\beta := (1 + \beta)^{-1} \in (0, 1)$.
The last proportion is obtained by dividing the raw gradient by $\beta (1 + \beta) p_V$.
Since $\beta$ is constant, it can be absorbed into the learning rate, but $p_V$ seems to cause a bias in the convergence destination, as questioned in the previous study \citep{kobayashi2022optimistic}.
However, we found that the Fisher information for $p(O|V)$ is given as $\beta^2 p_V (1 - p_V)^{-1}$, and dividing the raw gradient by $p_V$ can be interpreted as the sum of the raw and natural gradients (details are in Appendix~\ref{app:natural}), which is expected to converge to the same destination without bias \citep{landro2020mixing}.

As a practical problem, $p_V$ and $p_Q$ cannot be numerically computed since they include the unknown $\overline{R}$, the upper bound of the reward and return.
The previous study approximates the above gradient by assuming $\lambda_\beta \to 0$ (i.e. $\beta \to \infty$), resulting in the standard TD learning (by assuming $Q(s,a) \simeq r + \gamma V(s^\prime)$).

In addition to the value function, the policy $\pi$ (more precisely, its parameters $\phi$) is also optimized through the following minimization problem.
\begin{align}
    &\min_\phi \mathbb{E}_{p_e}[
    \mathrm{KL}(\pi(a \mid s) \mid \pi(a \mid s, O=1))
    - \mathrm{KL}(\pi(a \mid s) \mid \pi(a \mid s, O=0))
    ]
    \nonumber \\
    =& \min_\phi \mathbb{E}_{p_e, \pi}\Biggl[\ln \cfrac{\pi(a \mid s, O=0)}{\pi(a \mid s, O=1)}\Biggr]
    \label{eq:prob_pol}
\end{align}
The gradient w.r.t. $\phi$, $g_\phi$, is also derived analytically.
\begin{align}
    g_\phi &=
    \mathbb{E}_{p_e,\pi}\Biggl [
    \cfrac{\nabla_\phi \pi(a \mid s)}{\pi(a \mid s)} \ln \cfrac{\pi(a \mid s, O=0)}{\pi(a \mid s, O=1)}
    \Biggr ]
    \nonumber \\
    &= \mathbb{E}_{p_e,\pi}\Biggl [
    - \nabla_\phi \ln \pi(a \mid s)
    \Biggl \{
    \beta (Q(s,a) - V(s)) + \ln \cfrac{1 - p_V}{1 - p_Q}
    \Biggr \}
    \Biggr ]
    \nonumber \\
    &\propto \mathbb{E}_{p_e, b}\Biggl [
    - \cfrac{\pi(a \mid s)}{b(a \mid s)} \nabla_\phi \ln \pi(a \mid s)
    \Biggl \{
    (1 - \lambda_\beta) (Q(s,a) - V(s))
    + \lambda_\beta \ln \cfrac{1 - p_V}{1 - p_Q}
    \Biggr \}
    \Biggr ]
    \label{eq:grad_pol}
\end{align}
where the last proportion is given by dividing the raw gradient by $(1 + \beta)$.
In addition, at the end, the importance sampling replaces $\pi$ for the expectation operation with the baseline policy $b$.
As well as the value function, the approximation of $\lambda_\beta \to 0$ makes this gradient computable, resulting in the standard policy gradient in actor-critic algorithms.

\section{Weber-Fechner law in TD learning}

\subsection{Numerical analysis with explicit upper bound}

\begin{figure*}[tb]
    \centering
    \includegraphics[keepaspectratio=true,width=0.96\linewidth]{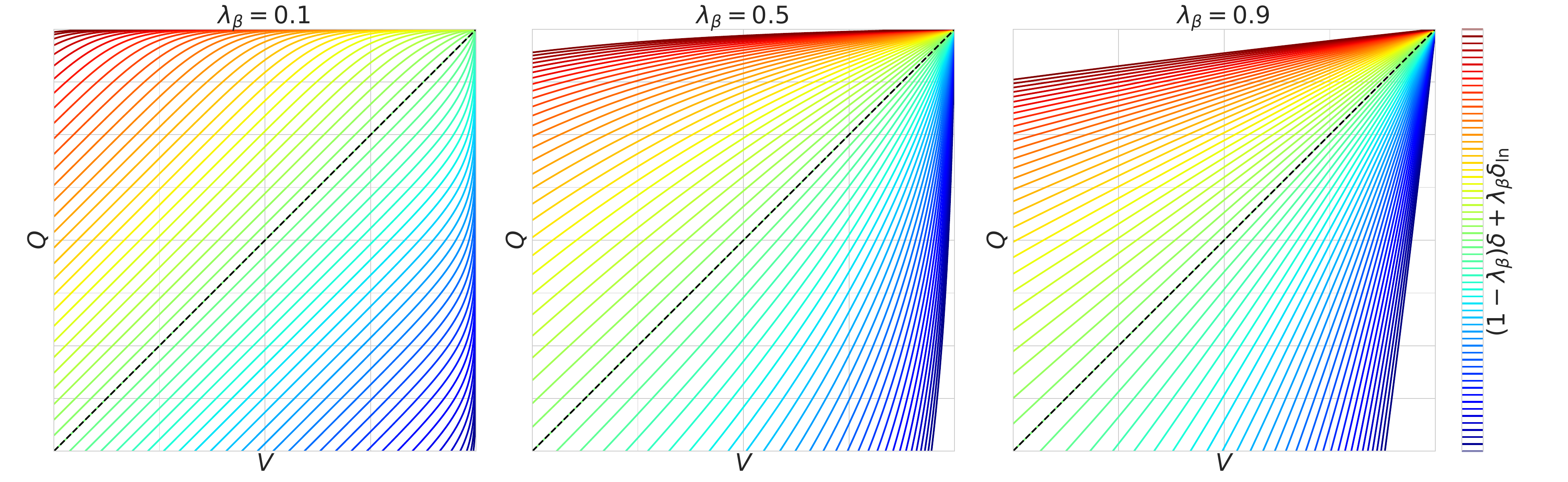}
    \caption{Effects of the nonlinear term $\delta_\mathrm{ln}$:
    when $\delta$ is dominant, the degrees of updates depicted by the contour lines are mostly equally spaced in parallel to the line of $V=Q$;
    when the influence of $\delta_\mathrm{ln}$ increases, the contour lines radiate out from the upper bound $\overline{R}$.
    }
    \label{fig:vis_effect_lambda}
\end{figure*}

The gradients to optimize the value and policy functions are derived in the eqs.~\eqref{eq:grad_val} and~\eqref{eq:grad_pol}, respectively.
However, as the upper bound of the reward function $\overline{R}$ is unknown and $p_V$ and $p_Q$ cannot be calculated numerically, it was necessary to exclude the uncomputable term by setting $\lambda_\beta \to 0$.
As a result, the previous study \citep{kobayashi2022optimistic} found the conventional update rule, where the gradients are weighted by $Q(s,a) - V(s) \simeq r + \gamma V(s^\prime) - V(s) = \delta$ (i.e. the TD error).
On the other hand, if the nonlinear term excluded (i.e. $\delta_\mathrm{ln}:=\ln (1 - p_V) - \ln(1 - p_Q)$) is computable, it is interesting how it affects the gradients, and that analysis is the main focus of this paper.

Therefore, we assume that $\overline{R}$ is known at once in this section.
With this assumption, the gradient including $\delta_\mathrm{ln}$ is analyzed.
First, we numerically visualize the gradient according to $\lambda_\beta \in (0, 1)$ and estimate the role of $\delta_\mathrm{ln}$, which will have a stronger influence when $\lambda_\beta$ increases (i.e. $\beta$ decreases).
For this purpose, $(1 - \lambda_\beta)\delta + \lambda_\beta \delta_\mathrm{ln}$ (i.e. the degree of updates) for $\lambda_\beta = \{0.1, 0.5, 0.9\}$ in the case $\mathcal{R} = (-1, 1)$ are illustrated in Fig.~\ref{fig:vis_effect_lambda}.

First, at $\lambda_\beta=0.1$, the contour lines representing the degree of updates are spaced equally and parallel to the line of $V=Q$.
This is mainly because $\delta$ is dominant, namely, the degree of updates is linearly proportional to the TD error.
Note that the behavior is slightly different for $V, Q \simeq \overline{R}$, because $\delta_\mathrm{ln}$ remains.
The remained $\delta_\mathrm{ln}$, however, easily converges to zero since the large $\beta$ ($=9$ in this case) makes $p_V$ and $p_Q$ converge to zero even with the small difference between $V, Q$ and $\overline{R}$ (i.e. the optimality is deterministic).

On the other hand, at $\lambda_\beta=0.9$, $\delta_\mathrm{ln}$ is dominant, resulting in that the contour lines extend radially from $\overline{R}$.
That is, when the value is close to $\overline{R}$, the update is significantly activated even with the small TD error; while when the value is far from $\overline{R}$, only the large TD error allows the update enough.
Unlike the case with $\lambda_\beta=0.1$, $\delta_\mathrm{ln}$ has a strong effect even when the value is far from $\overline{R}$ because $p_V$ and $p_Q$ with the small $\beta$ ($=1/9$ in this case) are changed around $1/2$ without converging to zero (i.e. the optimality is uncertain).

Finally, $\lambda_\beta=0.5$ yields an intermediate behavior between the above two characteristics.
That is, when the value is somewhat close to $\overline{R}$, the radial spread from $\overline{R}$ is observed due to the influence of $\delta_\mathrm{ln}$, and when it falls below a certain level, $\delta$ dominates and it switches to parallel contour lines.
However, it should be noted that this trend depends on the range of $\mathcal{R}$, so it is not always true for $\lambda_\beta=\{0.1, 0.5, 0.9\}$.

\subsection{Mathematical analysis with Taylor expansion}

We mathematically deepen the characteristic of $\delta_\mathrm{ln}$ found in the above numerical analysis.
Since it becomes apparent when $V$ and $Q$ are close to $\overline{R}$, Taylor expansion is applied to $p_V$ and $p_Q$ around $\overline{R}$.
\begin{align}
    p_V &= \sum_{n=0}^\infty \cfrac{\beta^n (V - \overline{R})^n}{n!} \simeq 1 + \beta (V - \overline{R})
    \\
    p_Q &= \sum_{n=0}^\infty \cfrac{\beta^n (Q - \overline{R})^n}{n!} \simeq 1 + \beta (Q - \overline{R})
\end{align}
Accordingly, $\delta_\mathrm{ln}$ is given as follows:
\begin{align}
    \delta_\mathrm{ln} = \ln \cfrac{1 - p_V}{1 - p_Q} \simeq - \ln \cfrac{\overline{R} - Q}{\overline{R} - V}
\end{align}

In that time, let's interpret $\overline{R} - V$ as the baseline of the stimulus strength, $\overline{R} - Q$ as the stimulus strength after the change (or $-\delta$ in $\overline{R} - Q \simeq \overline{R} - V - \delta$ as the change in stimulus strength), and $-\delta_\mathrm{ln}$ as the intensity of perception.
If this is the case, these are subject to \textit{Weber-Fechner law} (WFL).
In other words, how strongly the stimulus change $-\delta$ is perceived (i.e. $-\delta_\mathrm{ln}$) depends on the baseline of the stimulus strength $\overline{R} - V$: the smaller $\overline{R} - V$ is, the more acute the sensation becomes, and vice versa.
This is exactly the characteristic found in the right side of Fig.~\ref{fig:vis_effect_lambda}, indicating that the approximation by Taylor expansion is valid.

We then conclude the analysis that WFL is hidden even in the update rule of RL derived from control as inference.
WFL has also been found in areas closely related to brain functions such as neuron firing patterns \citep{scheler2017logarithmic} and cognition \citep{portugal2011weber}.
Therefore, it is not implausible to find it in RL, which is also attracting attention as a biological decision-making model \citep{dayan2002reward,doya2021canonical}.
This hypothesis would be supported by the fact that WFL is activated when the optimality is uncertain, which is consistent with the conditions faced by organisms \citep{parr2022active,tversky1974judgment}.

Furthremore, its learning availability and/or practical engineering value should be verified through numerical experiments in order to anticipate that WFL is a useful characteristic in RL and should be evolutionary left in organisms.
However, the above analysis was performed under the assumption that $\mathcal{R}$ is known, which is contrary to the general problem statement for RL.
In the next section, therefore, we propose a practical implementation that enables the computation of $\delta_\mathrm{ln}$ even when $\mathcal{R}$ is unknown in a biologically-plausible manner, followed by an experimental verification of the benefits of WFL in RL.

\section{Practical implementation}

\subsection{Introduction of reward-punishment framework}

First, we address the unknown $\mathcal{R}$ without giving prior knowledge of the problem to be solved.
The requirements are i) the boundary of $\mathcal{R}$ to define the optimality and ii) the range of $\mathcal{R}$ to determine $\beta$ (or $\lambda_\beta$) for which WFL is valid.
A naive solution would be to empirically estimate the boundary $(\underline{R}, \overline{R})$.
The estimation of the range $(\overline{R} - \underline{R})$ does not need to be rigorous, so learning can proceed stably if it is updated slowly.
However, in order for $p_V$ and $p_Q$ to always satisfy the definition of probability, $\overline{R}$ might be prone to be changed frequently, causing instability in learning.
Moreover, the empirically-estimated $\overline{R}$ would be underestimated from its true value, not bringing out the full utilities of WFL.

\begin{figure}[tb]
    \centering
    \includegraphics[keepaspectratio=true,width=0.96\linewidth]{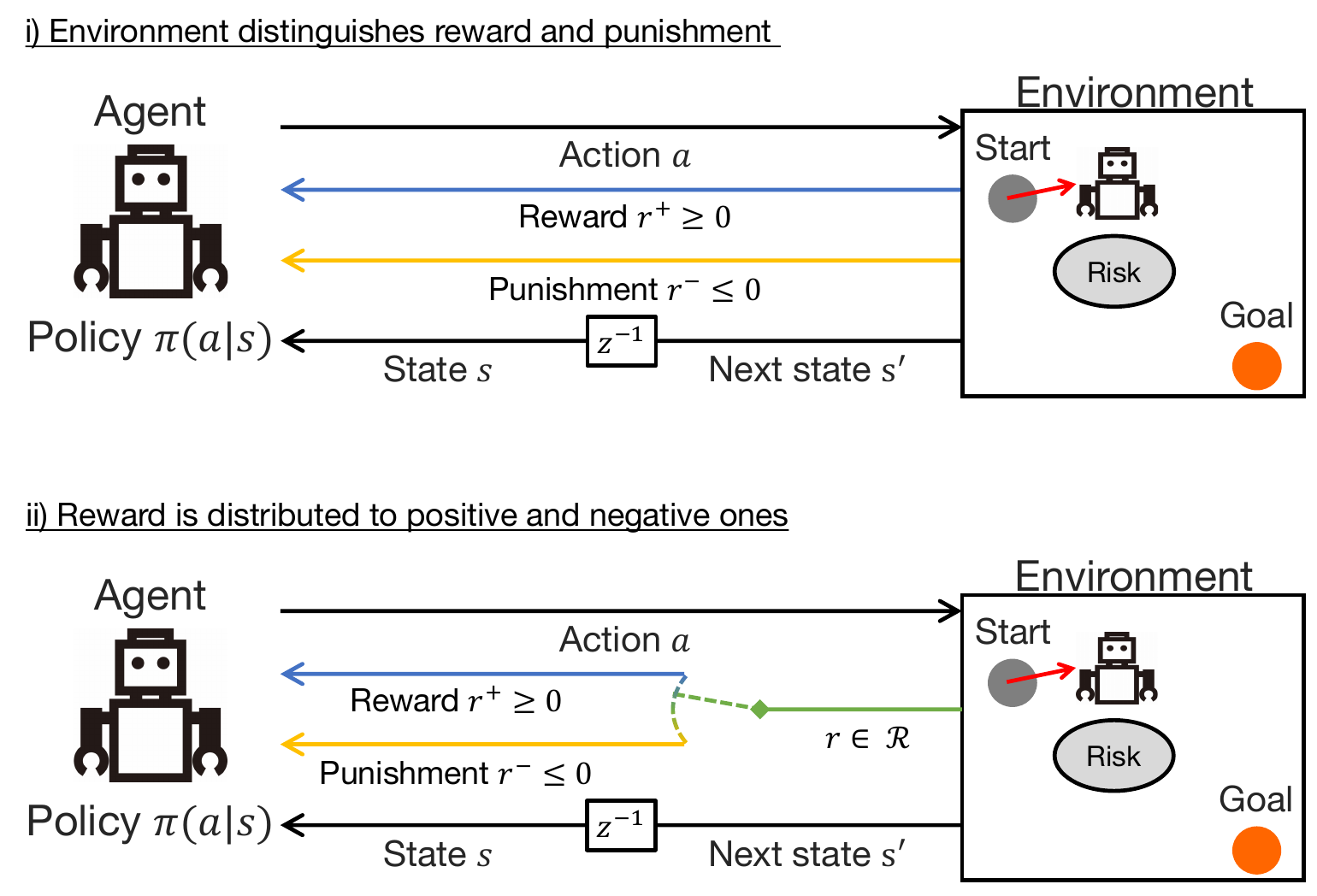}
    \caption{Reward-punishment framework:
    scalar rewards in a subset of real space are treated by distinguishing between positive and negative ones.
    }
    \label{fig:framework_rp}
\end{figure}

Therefore, in this paper, we introduce a more reliable solution, \textit{reward-punishment framework} \citep{kobayashi2019reward,wang2021modular}.
Although rewards are generally defined as scalar values that can be either positive or negative, this is a way to separate positive rewards $r^+ \in \mathcal{R}_+ \subseteq \mathbb{R}_+$ and negative rewards, i.e. punishments, $r^- \in \mathcal{R}_- \subseteq \mathbb{R}_-$.
This can be applied to any RL problem, either i) by having the environment output $r^{+,-}$ as in the experiments of this paper, or ii) by distributing $r \in \mathcal{R}$ from the environment as $r^+ = \max(r, 0)$ and $r^- = \min(r, 0)$ (see Fig.~\ref{fig:framework_rp}).

The reward-punishment framework learns the value and policy functions for $r^{+,-}$, respectively.
Specifically, the returns and the value functions for $r^{+,-}$ are first defined as follows:
\begin{align}
    \begin{split}
        R^{+,-}_t &= (1 - \gamma) \sum_{k=0}^\infty \gamma^k r^{+,-}_{t+k}
        \\
        V^{+,-}(s) &= \mathbb{E}_{p_\tau}[R^{+,-}_t \mid s_t = s]
        \\
        Q^{+,-}(s,a) &= \mathbb{E}_{p_\tau}[R^{+,-}_t \mid s_t = s, a_t=a]
    \end{split}
\end{align}
The policies $\pi^{+,-}$, which attempt to maximize them separately, are also introduced.

Here, since only one action can be passed to the environment even if the agent has two policies, it is necessary to synthesize them.
Following the previous study \citep{wang2021modular}, a mixture distribution with a mixing ratio based on the value function is designed.
\begin{align}
    \b(a \mid s) &= w \pi^+(a \mid s) + (1 - w) \pi^-(a \mid s)
    \\
    w &= \cfrac{e^{\beta_w V^+(s)}}{e^{\beta_w V^+(s)} + e^{-\beta_w V^-(s)}}
\end{align}
With this design, however, only one of the policy might be activated and the other might be ignored if the difference in the scales of $r^{+,-}$ is large.
To alleviate this issue, a policy regularization method, PPO-RPE \citep{kobayashi2023proximal}, for the density ratio $\pi^{+,-}/b$ with the importance sampling (see eq.~\eqref{eq:grad_pol}) is introduced in this paper.
As it yields $\pi^+ \simeq \pi^-$ via $b$, the past mixture distribution, transferring and sharing the acquired skills with each other.

Anyway, with the reward-punishment framework, the upper bound of $r^-$ is given to be zero, making $\delta_\mathrm{ln}$ for it computable.
On the other hand, $r^+$ only has the lower bound on zero, so $\delta_\mathrm{ln}$ for it remains uncomputable as it is.
In the next section, therefore, we derive $\delta_\mathrm{ln}$ utilizing this lower bound.

The range of rewards, $\sigma^{+,-}$, which is necessary for designing $\beta$ where WFL is effectively manifested, can be estimated from the empirical $r^{+,-}$.
However, the assumption when deriving eqs.~\eqref{eq:grad_val} and ~\eqref{eq:grad_pol} (i.e. $\beta$ is constant) is violated if $\sigma^{+,-}$ fluctuates too much.
In addition, since the experienced scale of $r^{-}$ is likely to gradually decrease, the approach to record the maximum scale is not suitable for this case.
From the above, $\sigma^{+,-}$ is estimated and used for the design of $\beta$ as follows:
\begin{align}
    \begin{split}
        \sigma^{+,-}_\mathrm{max} &\gets \max(\zeta \sigma^{+,-}_\mathrm{max}, |r^{+,-}|)
        \\
        \sigma^{+,-} &\gets \zeta \sigma^{+,-} + (1 - \zeta) \sigma^{+,-}_\mathrm{max}
        \\
        \beta^{+,-} &= \cfrac{\beta_0}{\sigma^{+,-}}
    \end{split}
    \label{eq:est_tde_scale}
\end{align}
where $\zeta \in (0, 1)$ denotes the gradualness of adaptation (generally, $\zeta$ is close to one) and $\beta_0 \in \mathbb{R}_+$ denotes the baseline.
This design allows $\beta$ to reflect the scale while limiting frequent fluctuations of $\beta$ by obtaining the recent maximum scale and updating to that value gradually.

\subsection{Inversion of definition of optimality with lower bound}

As mentioned above, although $r^- \leq 0$ can numerically compute eqs.~\eqref{eq:grad_val} and ~\eqref{eq:grad_pol} without any approximation, $r^+ \geq 0$ cannot do so since it only has the lower bound.
To solve this issue, another way of deriving the gradients with WFL is newly introduced, inspired by the previous study \citep{kobayashi2024drop}.
That is, the inversion of definition of optimality in eq.~\eqref{eq:prob_optim} is considered as the starting point.
\begin{align}
    \begin{split}
        p(O=0 \mid s) &= e^{-\beta (V(s) - \underline{R})} =: p_V
        \\
        p(O=0 \mid s, a) &= e^{-\beta (Q(s,a) - \underline{R})} =: p_Q
    \end{split}
    \label{eq:prob_optim_inv}
\end{align}
where the lower bound $\underline{R}$ of $\mathcal{R}$ is utilized for satisfying the definition of probability.
Note that the aliases $p_V$ and $p_Q$ are given for $p(O=0)$, unlike eq.~\eqref{eq:prob_optim}.

\begin{figure*}[tb]
    \centering
    \includegraphics[keepaspectratio=true,width=0.96\linewidth]{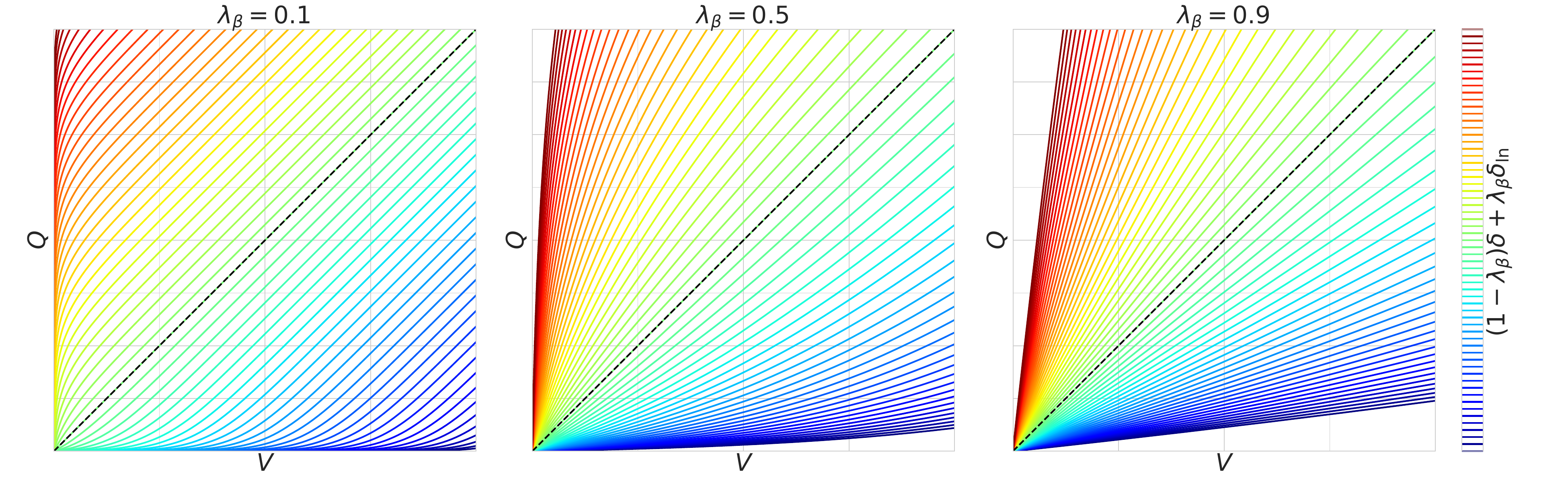}
    \caption{Weber-Fechner law with the lower bound:
    with the large $\lambda_\beta$, the contour lines become narrower with $V$ and $Q$ closer to their lower bound $\underline{R}$, and wider with $V$ and $Q$ farther from $\underline{R}$.
    }
    \label{fig:vis_effect_inversion}
\end{figure*}

As the previous study did not derive the gradients of eqs.~\eqref{eq:prob_val} and~\eqref{eq:prob_pol} with eq.~\eqref{eq:prob_optim_inv}, their derivations are described below.
First, $g_\theta$ for eq.~\eqref{eq:prob_val} is derived as follows:
\begin{align}
    g_\theta &=
    \mathbb{E}_{p_e, b}\Biggl [
    \nabla_\theta p_V \ln \cfrac{p_V}{p_Q}
    - \nabla_\theta p_V \ln \cfrac{1 - p_V}{1 - p_Q}
    + p_V \nabla_\theta \ln p_V
    + (1 - p_V) \nabla_\theta \ln (1 - p_V)
    \Biggr ]
    \nonumber \\
    &\propto \mathbb{E}_{p_e, b}\Biggl [
    - \nabla_\theta V(s)
    \Biggl \{
    (1 - \lambda_\beta) (Q(s,a) - V(s))
    - \lambda_\beta \ln \cfrac{1 - p_V}{1 - p_Q}
    \Biggr \}
    \Biggr ]
    \label{eq:grad_val_inv}
\end{align}
Except for the different definitions of $p_V$ and $p_Q$ and the sign reversal of the second term, it is similar to eq.~\eqref{eq:grad_val}.
Similarly, $g_\phi$ for eq.~\eqref{eq:prob_pol} is shown below.
\begin{align}
    g_\phi &=
    \mathbb{E}_{p_e,\pi}\Biggl [
    \cfrac{\nabla_\phi \pi(a \mid s)}{\pi(a \mid s)} \ln \cfrac{\pi(a \mid s, O=0)}{\pi(a \mid s, O=1)}
    \Biggr ]
    \nonumber \\
    &\propto \mathbb{E}_{p_e, b}\Biggl [
    - \cfrac{\pi(a \mid s)}{b(a \mid s)} \nabla_\phi \ln \pi(a \mid s)
    \Biggl \{
    (1 - \lambda_\beta) (Q(s,a) - V(s))
    - \lambda_\beta \ln \cfrac{1 - p_V}{1 - p_Q}
    \Biggr \}
    \Biggr ]
    \label{eq:grad_pol_inv}
\end{align}
where $\pi(a \mid s, O=0) = p_Q p_V^{-1} b(a \mid s)$ and $\pi(a \mid s, O=1) = (1 - p_Q) (1 - p_V)^{-1} b(a \mid s)$ are also redefined here.
This gradient is also the same format as eq.~\eqref{eq:grad_val_inv}, reversing the sign of second term, which has the different definitions of $p_V$ and $p_Q$.

Both have the same degree of updates multiplied by the gradients, and the first term coincides with the TD error $\delta$ as in the original.
The crucial second term appears to be different, but as depicted in Fig.~\ref{fig:vis_effect_inversion}, the contour lines extend radially from $\underline{R}$, as like the original.
In fact, if Taylor expansion around $\underline{R}$ is conducted to $p_V$ and $p_Q$, which are substituted for $\delta_\mathrm{ln} := - \ln (1 - p_V) + \ln (1 - p_Q)$, the same WFL is confirmed.
Thus, it is possible to compute the gradients with WFL even for $r^+$ where only $\underline{R} = 0$ is known.

\subsection{Expected utilities}

\begin{table*}[tb]
    \caption{Correspondence between WFL and the proposed update rule}
    \label{tab:corr}
    \centering
    \begin{tabular}{lccc}
        \hline\hline
        & WFL & Update rule for $r^+$ & Update rule for $r^-$
        \\
        \hline
        Formula & $p \propto \cfrac{S}{S_0}$ & $\delta_\mathrm{ln} \propto \ln \cfrac{Q}{V}$ & $- \delta_\mathrm{ln} \propto \ln \cfrac{-Q}{-V}$
        \\
        Baseline of stimulus strength & $S_0$ & $V$ & $-V$
        \\
        Stimulus strength after change & $S$ & $Q$ & $-Q$
        \\
        Change in stimulus strength & $S - S_0$ & $Q-V \simeq \delta$ & $-(Q-V) \simeq -\delta$
        \\
        Intensity of perception & $p$ & $\delta_\mathrm{ln}$ & $-\delta_\mathrm{ln}$
        \\
        \hline\hline
    \end{tabular}
\end{table*}

As described above, we have proposed a novel algorithm including the terms with WFL, which had been excluded in the previous study \citep{kobayashi2022optimistic} (and the standard RL algorithms) because they are computationally infeasible.
Table~\ref{tab:corr} summarizes the correspondence between WFL and the update rule in the proposed algorithm.
Note that since WFL is a law about the signal strength, the terms in punishments are converted for the punishment strength by reversing their signs.

Here, we summarize the basic utilities of WFL in this algorithm.
First, for rewards $r^+$, the updates of the value and policy functions are actively promoted at $V^+ \simeq 0$, while the updates are relatively suppressed over a certain level, $V^+ \gg 0$.
Conversely, for punishments $r^-$, the updates are slow under a certain level $V^- \ll 0$, but $V^- \simeq 0$ is pursued eventually.
It is known in gradient-based optimization that large gradients per update lead to a solution robust to small pertubations, while small gradients lead to one of the local solutions \citep{smith2018don,foret2021sharpness}.
Therefore, WFL in the proposed algorithm can also be interpreted as seeking a local solution for $r^+$ early and a global solution for $r^-$ steadily.
Note that, as shown in Figs.~\ref{fig:vis_effect_lambda} and~\ref{fig:vis_effect_inversion}, the utilities of WFL can be suppressed by adjusting $\lambda_\beta$ (or $\beta$) with the activation of the standard TD learning, which is not affected by the baseline stimulus strength (i.e. $|V|$).

\section{Numerical Verification}

\subsection{Toy problem}

\begin{figure}[tb]
    \centering
    \includegraphics[keepaspectratio=true,width=0.96\linewidth]{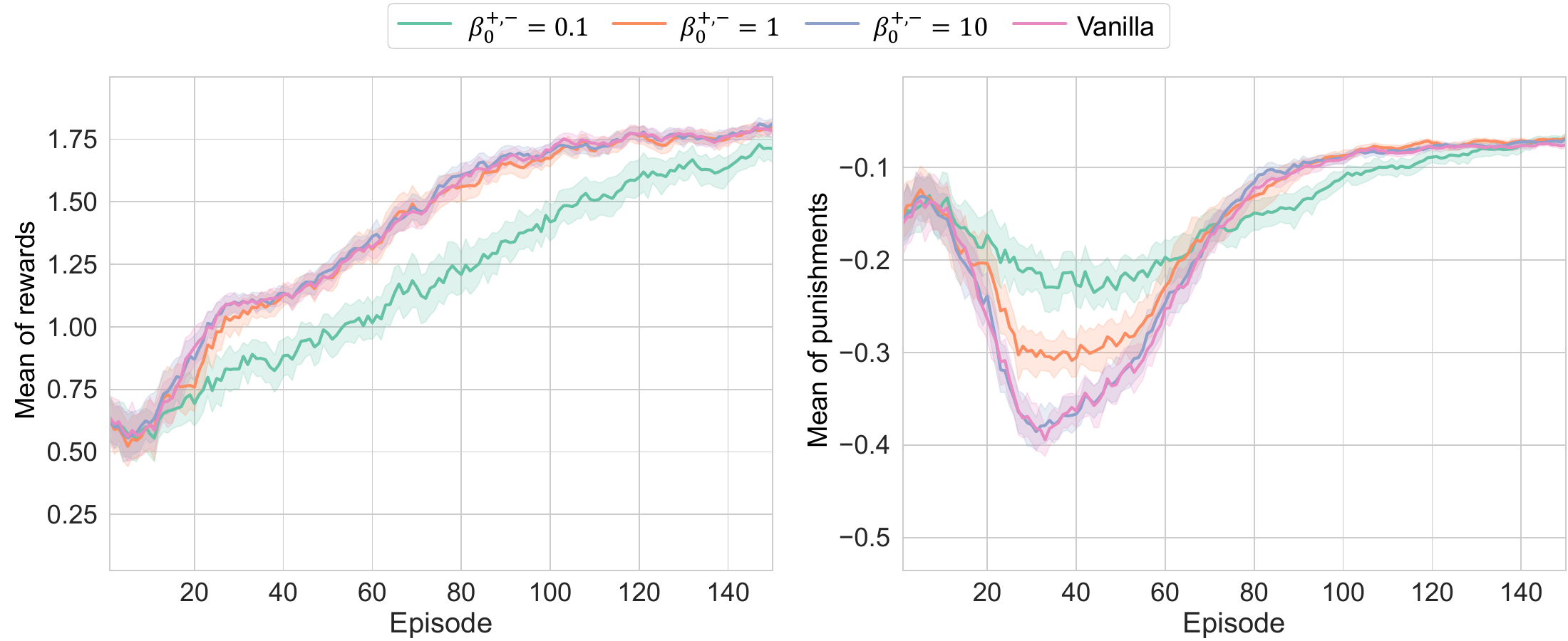}
    \caption{Learning results of \textit{Pendulum-v0} with different $\beta_0^{+,-}$:
    the upper and bottom curves depict the learning curves for episodic averages of $r^+$ and $r^-$, respectively;
    when $\beta_0^{+,-}$ was small, the pursuit of $r^+$ became slower and $r^-$ was preferentially suppressed to zero.
    }
    \label{fig:result_toy_beta}
\end{figure}

\begin{figure}[tb]
    \centering
    \includegraphics[keepaspectratio=true,width=0.96\linewidth]{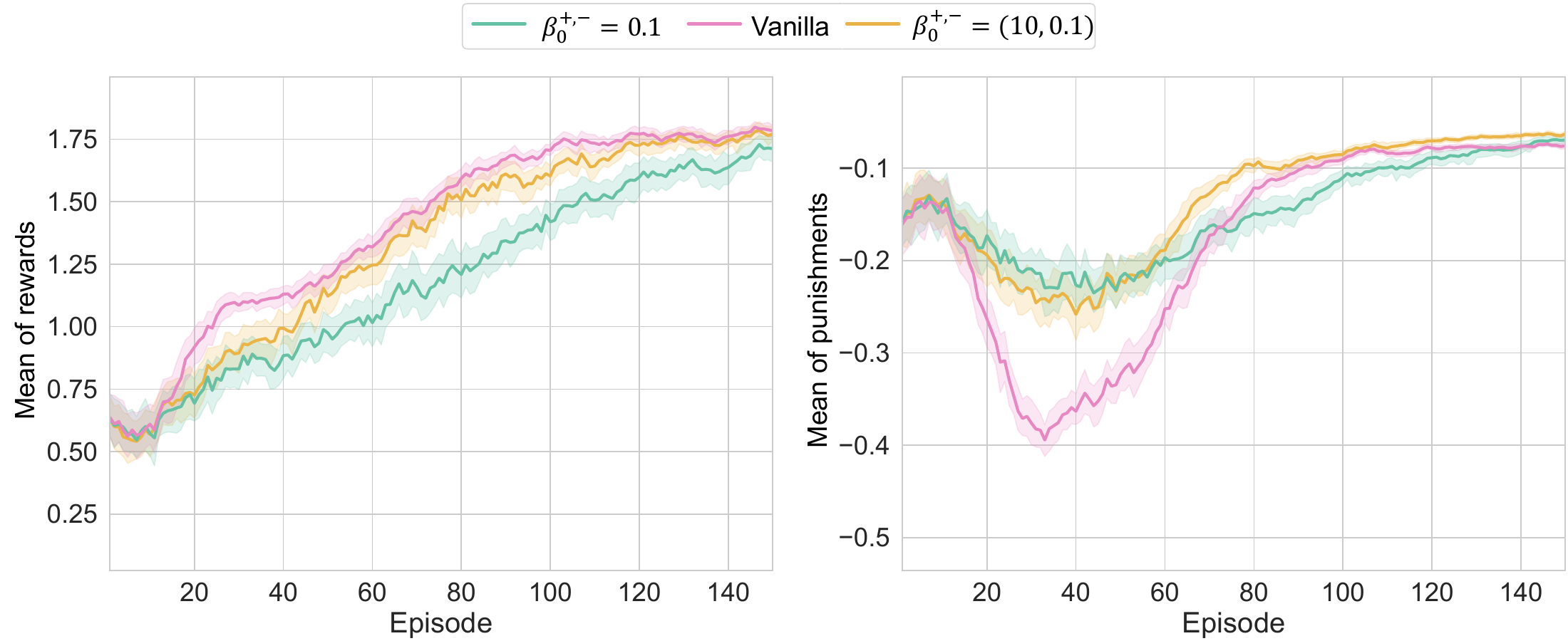}
    \caption{Learning results of \textit{Pendulum-v0} with the asymmetric $\beta_0^{+,-}$:
    thanks to the small $\beta_0^- = 0.1$, the deterioration of $r^-$ was restricted;
    although the exploration was more or less limited, the large $\beta_0^+ = 10$ enabled to maximize $r^+$ to the same level as the conventional method.
    }
    \label{fig:result_toy_sep}
\end{figure}

First, we investigate the feasibility of learning the optimal policy under the proposed algorithm with WFL, and the effects of WFL on the learning process and results.
As a toy problem, \textit{Pendulum-v0} implemented in OpenAI Gym is employed, while its reward function is redefined to fit the reward-punishment framework.
\begin{align}
    \begin{split}
        r^+ &= 1 + \cos q
        \\
        r^- &= - (0.1 |\dot{q}| + 0.001 |\tau|)(1 + \cos q)
    \end{split}
\end{align}
where $q$ denotes the pendulum angle, $\dot{q}$ denotes its angular velocity, and $\tau$ denotes the torque applied to the pendulum (i.e. action).
That is, the agent gets high rewards if the pendulum is close to upright, while it is punished when the pendulum is not stopped.
In addition, this punishment is stronger when the pendulum is close to upright.

With $\beta_0^{+,-} = \{0.1, 1, 10, \infty\}$ ($\infty$ means the conventional TD learning), four respective models are trained 50 times with different random seeds in order to achieve the statistical learning results in Fig.~\ref{fig:result_toy_beta}.
Note that the learning conditions including network architectures are summarized in Appendix~\ref{app:cond}.
As can be seen from the results, learning can proceed for any $\beta_0^{+,-}$ without collapse, meaning that RL is valid even if the term $\delta_\mathrm{ln}$ with WFL is used.
The increase of $\beta_0^{+,-}$ made the learning curves close to the conventional ones, as expected.
On the other hand, when $\beta_0^{+,-}$ becomes smaller, the unique behaviors were observed.
That is, for $\beta_0^{+,-} = 0.1$, the optimization with respect to $r^+$ was delayed while the temporary deterioration of $r^-$ was restricted.
This can be attributed to the fact that the optimization worked to reduce $r^-$ to zero as much as possible, while satisfying the maximization of $r^+$ at some level.
However, in the former case, it might be possible that the strong effort to reduce $r^-$ to zero suppressed the exploration, causing a delay in the discovery of the optimal solution for $r^+$.

Then, to take the advantages of both, the results of setting $\beta_0^+ = 10$ and $\beta_0^- = 0.1$ asymmetrically are depicted in Fig.~\ref{fig:result_toy_sep}.
Under this setting, WFL's efforts to reduce $r^-$ to zero remained and suppressed the temporary deterioration of $r^-$, while $r^+$ was successfully optimized without much delay.
In other words, the delay in learning about $r^+$ at $\beta_0^{+,-} = 0.1$ can be judged to be due to the characteristics of WFL.
Note that the delay in maximizing $r^+$ at around 20 episode is considered to be the effect of the suppression of exploration.

Anyway, WFL's utilities analyzed in this paper were confirmed as expected in the numerical verification.
In addition, as suggested in Fig.~\ref{fig:result_toy_sep}, the optimization behaviors for $r^{+,-}$ can be adjusted by setting $\beta_0^{+,-}$ separately.
However, we need to remark that the separation of $\beta_0^{+,-}$ does not mean that they function independently, since $r^{+,-}$ necessarily depend on each other in the learning process, as in the exploration suppression described above.

\subsection{Robotic task}

\begin{figure}[tb]
    \centering
    \includegraphics[keepaspectratio=true,width=0.96\linewidth]{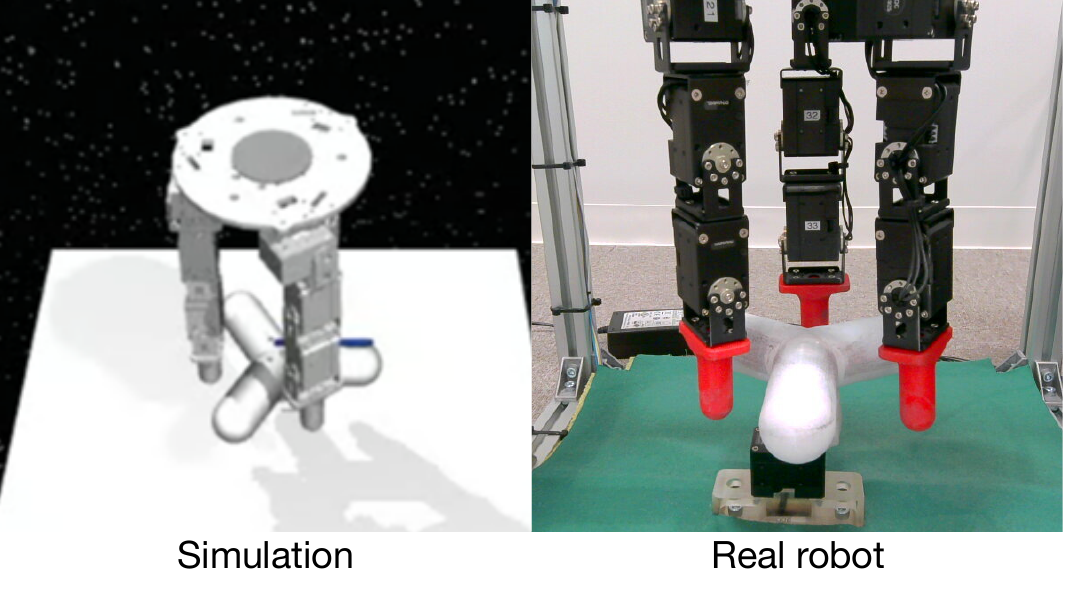}
    \caption{D'Claw task \citep{ahn2020robel}:
    it is simulated on Mujoco \citep{todorov2012mujoco}.
    }
    \label{fig:task_dclaw}
\end{figure}

The above numerical verification showed that the proposed method with WFL can optimize the policies with its expected learning characteristics.
Based on this finding, we additionally demonstrate that the proposed method can be useful in more practical robotic tasks.
This paper focuses on the D'Claw task in ROBEL \citep{ahn2020robel}, in which three 3-DOF robot fingers manipulate a valve (see Fig.~\ref{fig:task_dclaw}).
Note that the code for this system is not the original one, but a modified version in the literature \citep{yamanokuchi2022randomized}.
Its state space consists of the angles and angular velocities of the finger joints and the angle and angular velocity of the valve (in total, 22 dimensions), while the action space consists of 9 dimensions of angular changes of the finger joints.

\subsubsection{Simulations}

\begin{figure}[tb]
    \centering
    \includegraphics[keepaspectratio=true,width=0.96\linewidth]{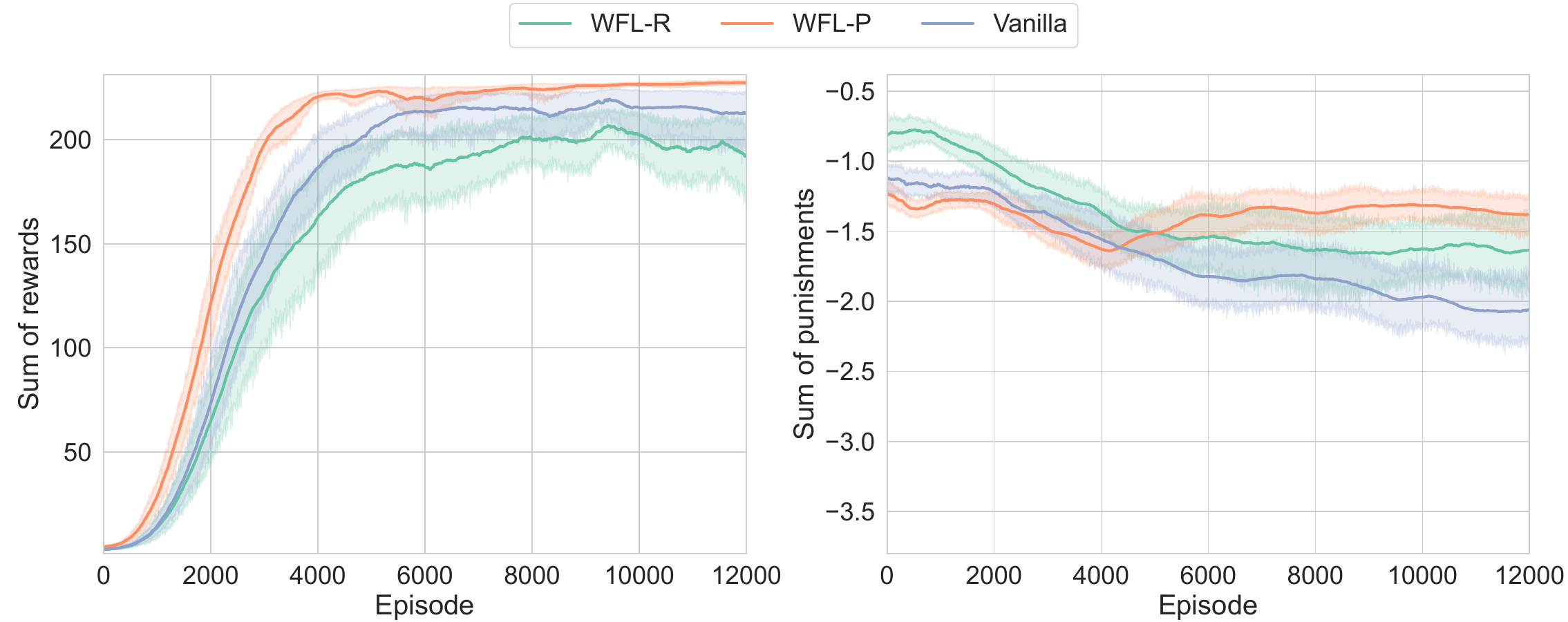}
    \caption{Learning results of D'Claw simulation:
    $\beta_0^{+,-} = (0.5, \infty), (\infty, 0.5)$ are labelled as \textit{WFL-R} and \textit{WFL-P}, respectively;
    the proposed method with WFL (especially, WFL-P) improved the maximization of rewards while reducing punishments.
    }
    \label{fig:result_sim}
\end{figure}

First, the simulations confirm that the behaviors when WFL is activated for $r^{+,-}$ individually are consistent with the toy problem.
Since $\beta_0^{+,-} = 0.1$ was too extreme and $\beta_0^{+,-} = 1$ was insufficient for WFL effects, $\beta_0^{+,-} = 0.5$ is adopted to activate WFL from here as \textit{WFL-R/P}.

The reward function is defined as follows:
\begin{align}
    \begin{split}
        r^+ &= q_v \mathbb{I}_{q_v > 0}
        \\
        r^- &= - 0.01 \| q_j + \Delta q_j \|^2_2
    \end{split}
\end{align}
where $q_v$ and $q_j$ denote the valve angle and the joint angles, respectively, and $\Delta q_j$ denotes the angular changes of joints (i.e. action).
That is, the goal is to turn the valve as much as possible while keeping the fingers in the initial posture to some extent.
Note that the sign of $q_v$ reverses after one turn, so the actual goal is to stop just before one turn.

The learning results of each condition with 20 different random seeds are shown in Fig.~\ref{fig:result_sim}.
Note that because the punishments were very small, unlike the toy problem, we plotted the sum of rewards/punishments per episode rather than their mean.
First, the pursuit of $r^+$ was slowed down in \textit{WFL-R} with WFL for $r^+$.
As its side effect, $r^-$ always outperformed the conventional method due to the reduced robot motion.
On the other hand, \textit{WFL-P} with WFL for $r^-$ showed improvement from around 4000 episode in pursue of reducing punishments, finally obtaining the best.
In addition, probably because the range of motion of each finger joint was maintained as its side effect, the speed of improvement of $r^+$ was significantly increased.
Thus, it was suggested that the appropriate addition of WFL, including its side effects, can improve learning performance in practical robotic tasks.

\subsubsection{Real-robot experiments}

\begin{figure}[tb]
    \centering
    \includegraphics[keepaspectratio=true,width=0.96\linewidth]{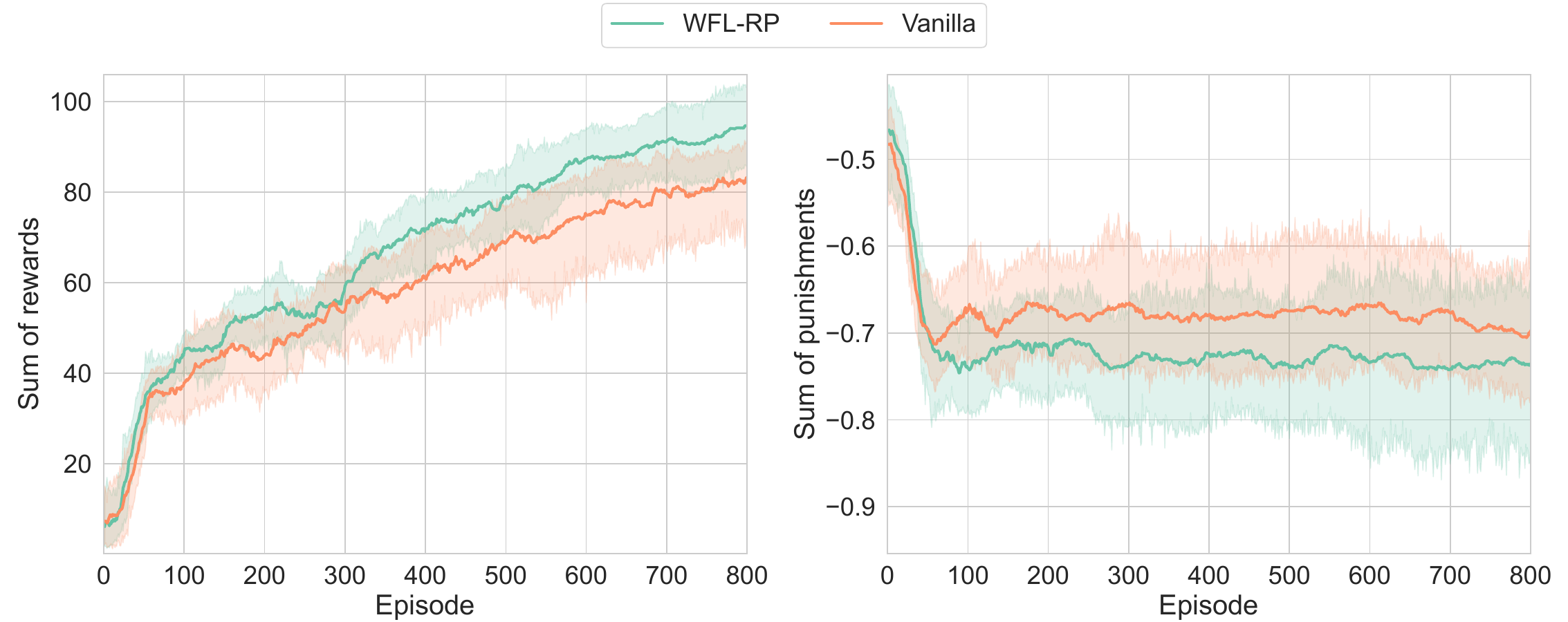}
    \caption{Learning results of real-world D'Claw:
    $\beta_0^{+,-} = (0.5, 0.5)$ is labelled as \textit{WFL-RP};
    accelerating the increase of a small reward was confirmed, although the increase in punishment by the side effect of the behavior for rotating the valve was not suppressed.
    }
    \label{fig:result_real_learn}
\end{figure}

\begin{figure}[tb]
    \centering
    \includegraphics[keepaspectratio=true,width=0.96\linewidth]{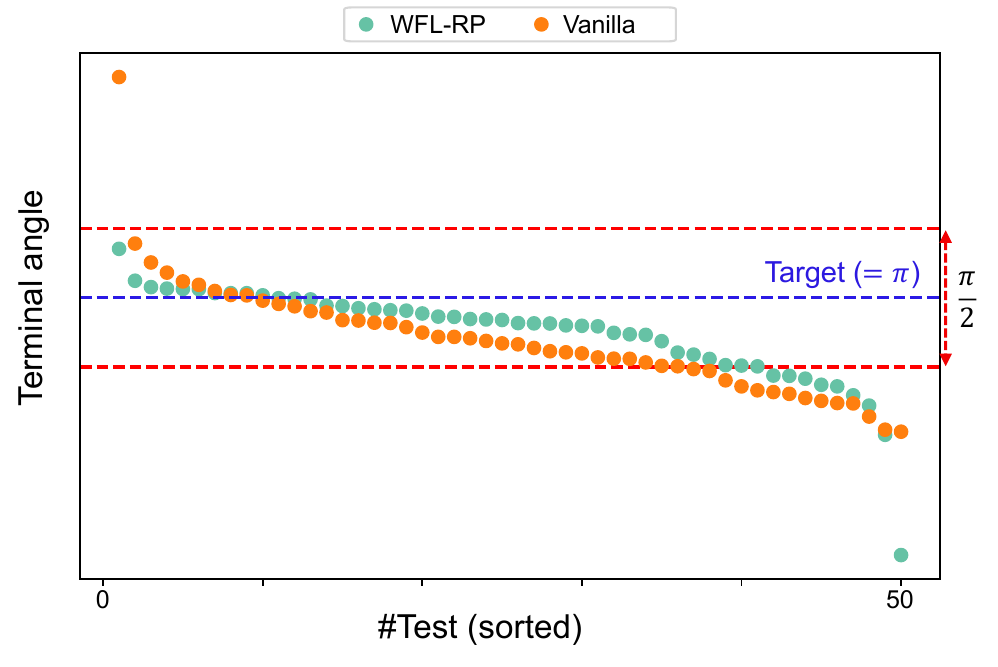}
    \caption{Terminal valve angles with the post-learning policies:
    10 episodes were performed in each of the five models, and the (sorted) valve angles at the end of the episodes were plotted;
    \textit{WFL-RP} tended to be closer to the target angle than the conventional method, and it also outperformed in the success rate where the error was within $\pi/4$.
    }
    \label{fig:result_real_test}
\end{figure}

Next, we demonstrate how WFL works in learning on the real world.
For simplicity, WFLs for $r^{+,-}$ are both activated simulutaneously and compared with the conventional method.

Since the real-robot valve angle has the different domain from the simulation one, and the angle jumps to $\pi \to -\pi$ in a half turn, the reward function is modified accordingly.
\begin{align}
    \begin{split}
        r^+ &= |q_v| \mathbb{I}_{q_v > 0 \lor q_v < -\frac{3}{4}\pi}
        \\
        r^- &= - 0.01 \| q_j + \Delta q_j \|^2_2
    \end{split}
\end{align}
In other words, the goal is to stop the valve half a turn while allowing some overshoot.
Note that, as the other differences from the above simulations, the ranges of motion and actions (i.e. the exploration capability) are restricted to avoid hardware malfunction.

First, the learning results with five trials are shown in Fig.~\ref{fig:result_real_learn}.
It was confirmed in $r^+$ that the proposed method always outperformed the conventional method.
This is probably because the proposed method preferentially learned a small number of motion samples that rotated the valve forward, which were rarely obtained by chance with the limited exploration capability.
Instead, $r^-$ of the proposed method was slightly lower than that of the conventional method, probably because the side effect of the behavior to rotate the valve was larger than the behavior to keep $r^-$ at zero.
Another posibility should also be noted that the regularization of $\pi^+ \simeq \pi^-$ was added, but the large difference in scale between $r^+$ and $r^-$ may have prevented it from functioning satisfactorily, and the policy to pursue $r^+$ may have been prioritized.

Next, task accomplishment, which cannot be evaluated from $r^{+,-}$ alone, is evaluated by the terminal valve angle with the post-learning policies.
The five post-learning policies for each condition tested 10 episodes, resulting in Fig.~\ref{fig:result_real_test}.
As expected from the learning curve of $r^+$, the proposed method produced more results closer to the target angle, $\theta=\pi$.
Moreover, when the range of $\pm\pi/4$ from the target angle is considered as the success, the proposed method showed $41/50$ (i.e. $82\%$), whereas the conventional method showed $35/50$ (i.e. $70\%$).

\section{Conclusion and discussion}

In this paper, we revealed a novel nonlinearity in TD learning, WFL, which explains the relationship between stimuli and perception of organisms, in the update rule of the value and policy functions in RL.
Without loss of generality, it was implemented as a novel biologically-plausible RL algorithm on the reward-punishment framework.
We showed that the proposed method can be expected to explore a local solution to maximize rewards as early as possible, while gradually aiming for a global solution to minimize punishments.
Numerical verification indicated that the proposed method does not collapse RL and provides the characteristics coming from WFL.
The proposed method was also useful for robot control, and it outperformed the conventional method in the valve-turning task using D'Claw.

As shown above, although WFL in TD learning confirmed in this study can be shown to produce more desirable learning processes and outcomes, the fact that WFL is more likely to be affected by reward design than the conventional TD learning is still open to investigation.
Basically, objectives given as punishments $r^-$ should have a high priority for achievement, and those with rewards $r^+$ should be regarded as value-added.
However, prioritization among multiple objectives is often given as weights, which may lead to overlapping roles among multiple parameters and make it difficult to understand.
The complexity is further increased by the fact that the impact of WFL can be adjusted by $\lambda_\beta$ (or $\beta$).
Therefore, learving the design of such priorities to RL users and/or task designers may be an obstacle to real-world applications.
To alleviate this issue, further research on the design theory of reward functions suitable for this algorithm and/or the automation of assignment to $r^{+,-}$ (and tuning of hyperparameters) based on user preferences would increase the practical value of this algorithm.

On the other hand, WFL found in TD learning originally explains the relationship between stimuli and perception in organisms.
Considering that RL is also employed as decision-making models for organisms, and that the relationship between TD errors and brain activities have actually been verified, it is possible that WFL in TD learning may be latent in our brain activities.
Therefore, it would be important to verify the existence or absense of WFL by using this algorithm for the analysis of brain-activity data.
Moreover, the feedback from the found insights may be able to elaborate our algorithm.

\appendix
\section*{Appendix}

\section{Natural gradient for $\theta$}
\label{app:natural}

The Fisher information for $p(O|V)$, $\mathcal{I}(V)$, is derived as follows:
\begin{align}
    \mathcal{I}(V) &= \mathbb{E}_{p(O|V)}\left[ \left( \frac{\partial}{\partial V} \ln p(O|V) \right)^2 \right]
    \nonumber \\
    &= p(O=1|V) \left( \frac{\partial}{\partial V} \ln p(O=1|V) \right)^2
    + p(O=0|V) \left( \frac{\partial}{\partial V} \ln p(O=0|V) \right)^2
    \nonumber \\
    &= \beta^2 p_V + (1 - p_V) \left( \frac{-\beta p_V}{1 - p_V} \right)^2
    \nonumber \\
    &= \beta^2 p_V \left( 1 + \frac{p_V}{1 - p_V} \right)
    \nonumber \\
    &= \beta^2 \frac{p_V}{1 - p_V}
\end{align}
where $p_V = p(O=1|V) = e^{\beta(V - \overline{R})}$.

The natural gradient is obtained by dividing the raw gradient by the Fisher information \citep{amari1998natural}.
As the raw gradient $g^\mathrm{raw}$ is represented as $- \{(1-\lambda_\beta) \delta + \lambda_\beta \delta_\mathrm{ln}\} \beta (1 + \beta) p_V \nabla_\theta V$, its natural gradient $g^\mathrm{nat}$ can be given as $- \{(1-\lambda_\beta) \delta + \lambda_\beta \delta_\mathrm{ln}\} \beta^{-1} (1 + \beta) (1 - p_V) \nabla_\theta V$.
By removing the constant coefficients related to $\beta$, their summation can be derived as $- \{(1-\lambda_\beta) \delta + \lambda_\beta \delta_\mathrm{ln}\} \nabla_\theta V$.

\section{Learning conditions}
\label{app:cond}

The value and policy functions for $r^{+,-}$ are independently approximated by a common network structure.
The structure has two fully connected layers as hidden layers, with 100 neurons for each, and ReLU function is employed as its activation function.
AdaTerm \citep{ilboudo2023adaterm} is employed to optimize the network parameters, and its learning rate is set to $1\times10^{-3}$ for the toy problem and $5\times10^{-4}$ for the robotic task.
The sample efficiency is improved with experience replay, although its buffer size is small (i.e. $1\times10^4$) to emphasize on-policyness.
Half of the stored experiences are randomly replayed at the end of each episode with 32 the batch size.
As tricks to stabilize learning, we introduce PPO-RPE \citep{kobayashi2023proximal} introduced in the main text, target networks with CAT-soft update \citep{kobayashi2024consolidated}, and regularization to make the functions smooth by L2C2 \citep{kobayashi2022l2c2}.
All of these remain the default hyperparameters.
Other hyperparameters are $\gamma=0.99$ for the discount rate and $\zeta=0.999$ for the empirical TD error scale estimation in eq.~\eqref{eq:est_tde_scale}.

\subsection*{Acknowledgments}

This work was supported by JSPS KAKENHI, Development and validation of a unified theory of prediction and action, Grant Number JP24H02176.

\bibliographystyle{apalike}
\bibliography{biblio}

\end{document}